# Seeds of Stereotypes: A Large-Scale Textual Analysis of Race and Gender Associations with Diseases in Online Sources

**Lasse Hyldig Hansen**[1,2], **Nikolaj Andersen** [1,2], **Jack Gallifant** [2,3], **Liam G. McCoy** [4], **James K Stone** [5], **Nura Izath** [6], **Marcela Aguirre-Jerez** [7], **Danielle S Bitterman** [8,9,10], **Judy Gichoya** [11], **Leo Anthony Celi** [2,12,13]

[1] Cognitive Science, Aarhus University, Jens Chr. Skou 2, 8000 Aarhus, Denmark
[2] Laboratory for Computational Physiology, Massachusetts Institute of Technology, Cambridge, Massachusetts, United States of America
[3] Department of Critical Care, Guy's & St Thomas' NHS Trust, London, United Kingdom
[4] Division of Neurology, University of Alberta, AB, Canada
[5] University of Manitoba Max Rady College of Medicine, Winnipeg, MB, Canada
[6] Faculty of Computing, Mbarara University of Science and Technology, Mbarara, Uganda.
[7] Digital Health Department, Fundacion Arturo Lopez Perez, Santiago, Chile
[8] Artificial Intelligence in Medicine (AIM) Program, Mass General Brigham, Harvard Medical School, Boston, Massachusetts, United States of America
[9] Department of Radiation Oncology, Brigham and Women's Hospital/Dana-Farber Cancer Institute, Boston, Massachusetts, United States of America
[10] Computational Health Informatics Program, Boston Children's Hospital, Harvard Medical School, Boston, Massachusetts, United States of America
[11] Department of Radiology, Emory University School of Medicine, Georgia, United States of America
[12] Division of Pulmonary, Critical Care, and Sleep Medicine, Beth Israel Deaconess Medical Center, Boston, Massachusetts, United States of America
[13] Department of Biostatistics, Harvard T.H. Chan School of Public Health, Boston, Massachusetts, United States of America

**Summary:**

*Background*

Advancements in Large Language Models (LLMs) hold transformative potential in healthcare, however, recent work has raised concern about the tendency of these models to produce outputs that display racial or gender biases. Although training data is a likely source of such biases, exploration of disease and demographic associations in text data at scale has been limited.

*Methods*

We conducted a large-scale textual analysis using a dataset comprising diverse web sources, including Arxiv, Wikipedia, and Common Crawl. The study analyzed the context in which various diseases are discussed alongside markers of race and gender. Given that LLMs are pre-trained on similar datasets, this approach allowed us to examine the potential biases that LLMs may learn and internalize. We compared these findings with actual demographic disease prevalence as well as GPT-4 outputs in order to evaluate the extent of bias representation.

*Results*

Our findings indicate that demographic terms are disproportionately associated with specific disease concepts in online texts. gender terms are prominently associated with disease concepts, while racial terms are much less frequently associated. We find widespread disparities in the associations of specific racial and gender terms with the 18 diseases analyzed. Most prominently, we see an overall significant overrepresentation of Black race mentions in comparison to population proportions.

*Conclusions*

Our results highlight the need for critical examination and transparent reporting of biases in LLM pretraining datasets. Our study suggests the need to develop mitigation strategies to counteract the influence of biased training data in LLMs, particularly in sensitive domains such as healthcare.

# Introduction

The general purpose ability of modern large language models (LLMs) has enabled broad applications across tasks and domains, facilitated by unprecedented growth in the availability of textual data and advancements in computing technologies. These LLMs undergo training on extensive collections of web-based data, sourced from comprehensive repositories like the Common Crawl, an open-source project that archives vast amounts of web data, and resources such as Wikipedia. Domain-specific models also rely on such approaches, typically training only the final layers of models on the target domain data whilst maintaining the remainder of the model as was developed during pretraining on the broad corpora [1]. These approaches have been extremely successful on current benchmarks with tools now entering the healthcare system on a wide array of tasks such as summarisation, clinical quality assurance (QA), and aiding reporting [2], [3], [4].

However, alongside this potential, concerns have arisen regarding the impacts of LLMs on health equity [5]. Recent studies have identified the presence of substantial biases in the outputs of LLMs both broadly [6], and when applied specifically to health-related tasks. Notably, recent work from Zack et al. [7] found the prominent model Generative Pre-trained Transformer 4 (GPT-4) to demonstrate race, ethnicity, and gender bias in several domains, including generating patient vignettes, recommending further testing, and diagnosing diseases. These findings underscore the need for a deeper investigation into the origins and nature of these biases to provide greater clarification on identifiability and potential for mitigation.

Training data provide the source of model representations [8], and the use of large-scale internet data allows for an incredible breadth of information to be incorporated; however, this also includes learning from the biases and stereotypes known to be contained in this data [9], [10]. These dataset characteristics have a significant influence on the output of LLMs, but despite this, detailed evaluation of the specific contents of this data at scale has been limited [8], [11]. Furthermore, there are limited methods available that provide quantitative measures of bias in training data linking these to downstream model behavior. It is also worth noting that the original training data is rarely disclosed.

This study introduces a method to quantify bias in training datasets by analyzing the associations between specific diseases and demographic variables within datasets frequently used in LLM development. We then assess the strength of these relationships, comparing disease and demographic associations in the training data with biases documented in [7]. This analysis compares these associations with GPT-4 outputs and actual disease prevalence across the United States.

## Methods
## Data Collection and Processing:

The primary dataset, referred to as "RedPajama-Data-1T" [12] is a curated collection created to mirror the known pre-training data of the Large Language Model Meta AI (LLaMA-1) [13]. This diverse dataset consists of six open-source text corpora: arXiv, Books, C4, GitHub, StackExchange, and Wikipedia (English), as summarized in ***Supplementary Table 1***. The dataset has been used to develop many prominent open-source LLMs and likely overlaps significantly with the data used to train large proprietary LLMs [13].

We systematically obtained text data in JSON Lines format from individual corpora within the RedPajama-Data-1T dataset by directly downloading files through standard web retrieval commands. Our preprocessing efforts focused on refining the dataset to isolate texts directly relevant to our research focus, namely examining textual references to diseases within specific contexts of demographic subgroups. We crafted regex (regular expression) patterns tailored to identify mentions within the texts of diseases and demographic keywords. To mirror the structure of Zack et al., we evaluate the co-occurrence by race/ethnicity (Black, White, Asian, or Hispanic) and gender (male or female). While gender is not binary, the limited real-world data for the prevalence of diseases among non-binary genders and the lack of available proportions from the aforementioned paper mean that only males and females are reported in this study. For simplicity, from hereon, the demographic categories are named race and gender.

### *Data Cleaning and Metadata Analysis:*

To account for the widespread use of LaTeX formatting within academic components of our dataset, we implemented a cleaning step to remove all LaTeX commands from the texts. Following this, we systematically analyzed each line of the JSONL files, parsing each JSON object to extract the text and any associated metadata available for each dataset. Texts containing references to the specified medical conditions, race or gender terms were retained for analysis, and pertinent metadata such as publication date, authorship, and source were recorded alongside a tally of the occurrences of the targeted keywords.

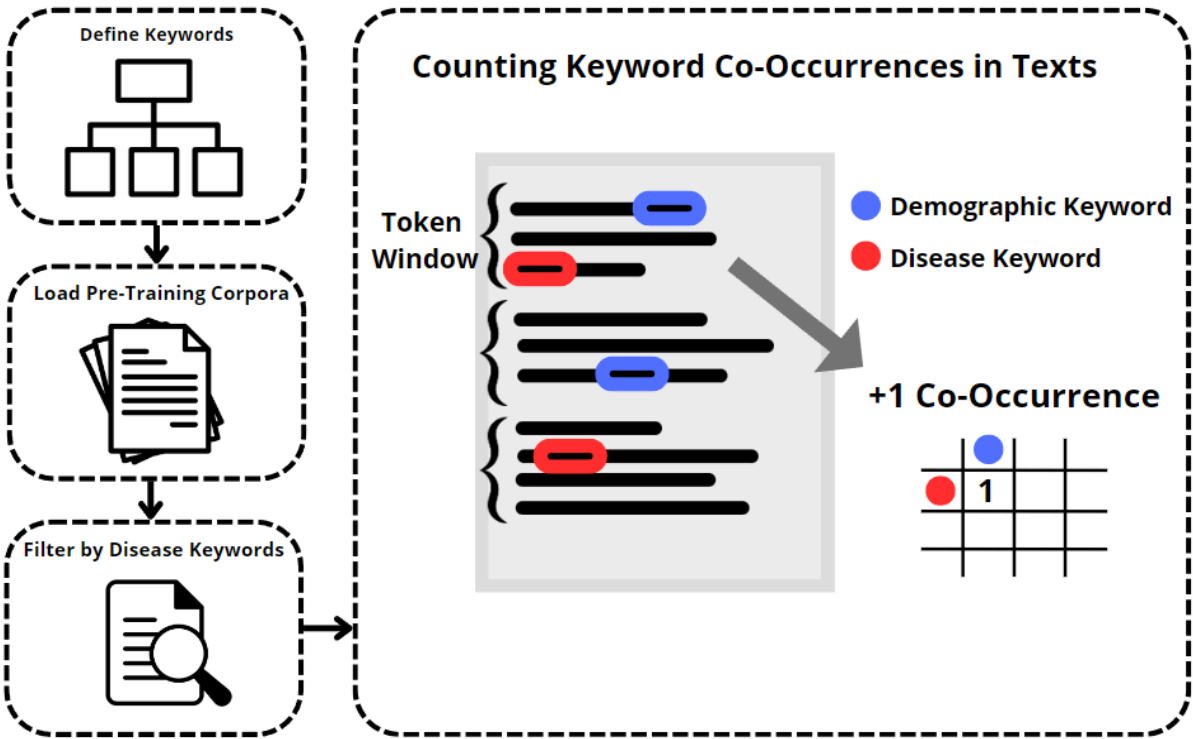

**Figure 1:** Summary of Data Processing Pipeline and Co-Occurrence Analysis

***Target Disease Selection and Keyword Definition:***

Keywords were selected based on the diseases reported in [7]. We manually created extensive lists of keywords for diseases, races, and genders, combining PubMed MeSH headings with additional terms selected by clinician co-authors (JG, LM, JS, LAC). The full dictionary of keywords and all code required for data analysis are available in the GitHub repository.

***Co-Occurrence Analysis:***

In our study, co-occurrence refers to the simultaneous presence of a disease keyword and a race or gender keyword within a specified range of text, measured in whitespace delimited words (See Figure 1). For each disease, we calculated co-occurrence counts for each race- and gender-related keyword within specified token windows. These token windows—20, 100, 200, 500 words, and the entire document—represent the range of text surrounding a disease keyword in which we search for the presence of race or gender keywords. Different window sizes allow us to evaluate the proximity of disease mentions to demographic identifiers, capturing both immediate and broader contextual relationships.

For example, in a 100-token window, we count a co-occurrence if a race or gender keyword appears within 50 words before or after a disease keyword in the text. These co-occurrence counts are then aggregated across all instances in the dataset to create a co-occurrence matrix for the 18 diseases, six races, and the two explored genders. This matrix provides a comprehensive overview of how frequently each disease is mentioned in proximity to each demographic identifier across the entire dataset.

*Analytical Approach - Representational Differences:*

We aggregated the counts of mentions associated with each keyword for each medical condition within the racial and gender datasets. These counts were used to calculate the percentage representation of each racial or gender keyword's mentions relative to the overall mentions of the disease.

$$Representation\ \% = \left(\frac{Count\ of\ Demographic\ Keyword\ Within\ Medical\ Context\ Window}{Total\ Count\ of\ Medical\ Context\ Windows}\right) * 100$$

Our analysis included the total disease counts as a separate keyword to capture instances where a medical term is mentioned without any immediate racial or gender context. We calculated proportions of disease mentions alongside demographic identifiers within all context windows, but ultimately focused on 100-word windows for the main analysis. We further segmented the data by source to explore nuances in how different text corpora represent the interplay between medical terms and demographic identifiers.

*Analytical Approach - Distribution vs. Real-World Prevalence:*

Data from [7] was used to integrate real-world disease prevalence data from a comprehensive literature review. This enabled a comparison of the frequency of disease mentions in relation to specific demographic groups in the text with the actual prevalence rates of these diseases among the same groups in the United States population, as well as with the racial frequency in GPT-4 outputs. Following their methods, this analysis was restricted to four racial categories from the initial six included (removing Pacific Islander and Native American). We used representation percentages for diseases within a 100-word window size as a proxy for the emphasis or focus that different diseases receive in relation to various demographic groups within the dataset.

*Context Window Size Influence on Representational Percentages:*

We calculated the representational percentage for each window size, quantifying the normalized frequency of a disease's co-occurrence with race or gender terms within the text. The percentages were calculated as the average over all data sources within the dataset. This analysis allowed us to understand if changes in the proximity of contextual analysis can influence the perceived association between diseases and demographic categories.

## Results

Analyzing the RedPyjama T1 dataset (1.2 Trillion Tokens), we find a total of 2.463.240.125 disease mentions of our target diseases, summing across all windows. As demonstrated in **Figure 2**, only a minority (43.27%, N = 8.630.598) of disease mentions are associated with gender descriptors at the 100-word window, and an even smaller proportion (8.23%, N = 1.023.803) are associated with race terms. These patterns vary somewhat between sources with, for example, Wikipedia (61.7%) and books (64.0%) having both a higher rate of gender co-occurrences than more technical sources such as ArXiv (4.5%) or Github (4.15%) [**Appendix Figure 2**].

Among windows where co-occurrences occur (Figure 3 and Table 1), we observed a significant overrepresentation of terms associated with Black racial groups and, to a lesser extent, terms related to Native American and Pacific Islander groups, as well as male gender terms. Concurrently, White and Hispanic racial terms are underrepresented, as well as female gender. Terms related to Asian groups most closely align with their actual population percentages. These results are consistent across window lengths [Appendix Figure 1], with some variation at the level of full documents.

| Category | Race/Ethnicity | | | | | | Gender | |
| --- | --- | --- | --- | --- | --- | --- | --- | --- |
| | White | Black | Asian | Hispanic | Native American | Pacific Islander | Female | Male |
| **Disease Co-occurrence** | | | | | | | | |
| Overall Proportion of Disease co-occurrences (100 word window) | 37.66% | 45.70% | 5.58% | 7.89% | 2.31% | 0.86% | 43.64% | 56.36% |
| **Population Proportion** | | | | | | | | |
| Population Proportion (2020 United States Census)‡ | 57.84%* | 12.05%* | 5.92%* | 18.73%† | 0.68%* | 0.19%* | 50.9% | 49.1% |

*Table 1: Overall Proportion of Demographic co-occurrences Relative to Population* *Non-Hispanic. †Denotes Hispanic or Latino origin of any race. ‡Proportions do not add up to 100% as US Census includes Some Other Race (0.51%) and Mixed Race / Multi-Racial (4.09%)*

We further compared our results to estimates of actual disease frequency for our target illnesses within the United States, as well as the outputs of GPT-4 when asked to generate a case of each disease **(Figure 4).** We find that the previously noted Black overrepresentation in the dataset is broad, observed in every disease noted except for hepatitis B. The degree of GPT-4 overrepresentation outstrips the degree of training data overrepresentation and is particularly stark for hypertension, prostate cancer, HIV/AIDS, lupus, and sarcoidosis. White markers are somewhat underrepresented broadly (most notably for prostate cancer), but overrepresented for tuberculosis.

Female markers are notably overrepresented in co-occurrences with HIV/AIDS, and underrepresented in co-occurrences with rheumatoid arthritis, relative to actual frequencies. Hispanic co-occurrences are notably underrepresented for nearly all diseases examined, with a particularly notable disparity in tuberculosis compared to both real-world prevalence and the GPT-4 estimate. Asian co-occurrences are relatively low throughout the sample, although slightly overrepresented for discussions of Hepatitis B and significantly overrepresented in the GPT-4 estimates. They are notably underrepresented in co-occurrences with tuberculosis, despite a high real-world prevalence and relatively high GPT-4 estimate.

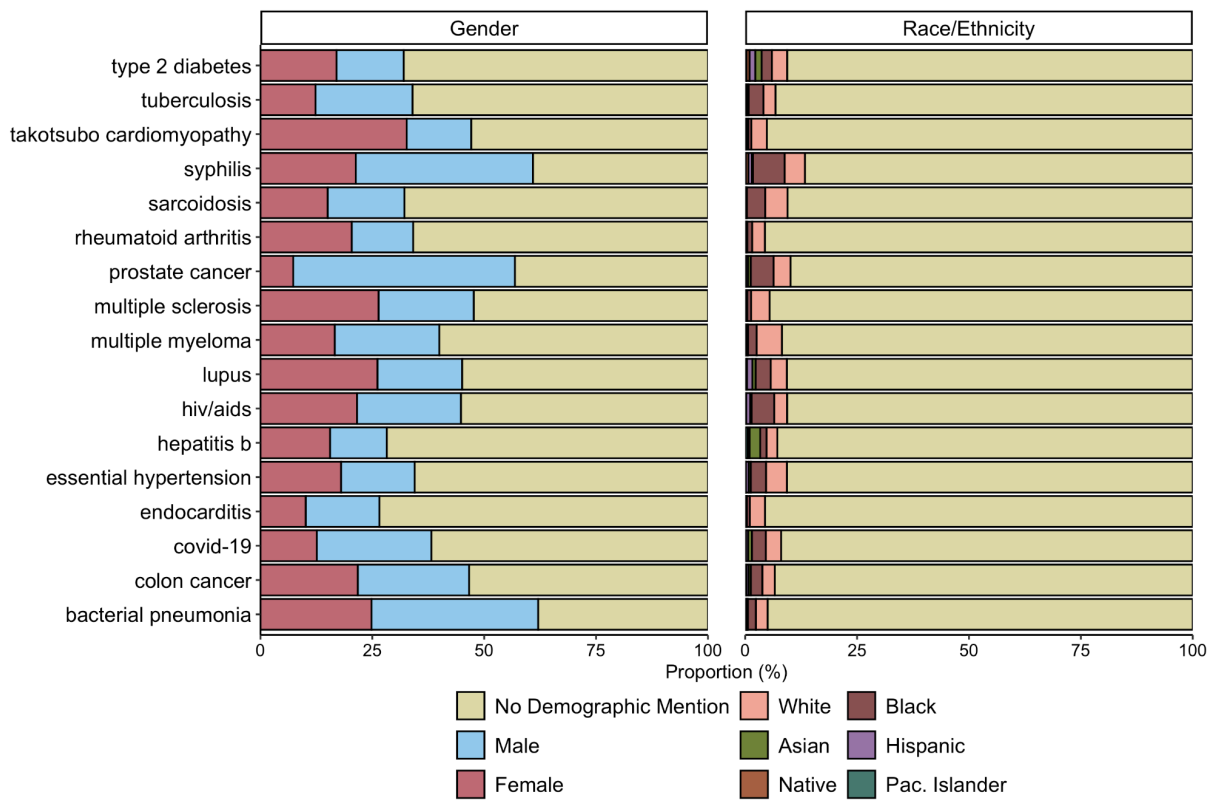

*Figure 2:* Proportional Disease Mentions with Demographic References within a 100-word Contextual Window. Panel A shows the gender-associated mentions of various diseases, with Panel B detailing the mentions in connection with different races. In both panels, yellow bars indicate the proportion of disease mentions occurring without any specific demographic context.

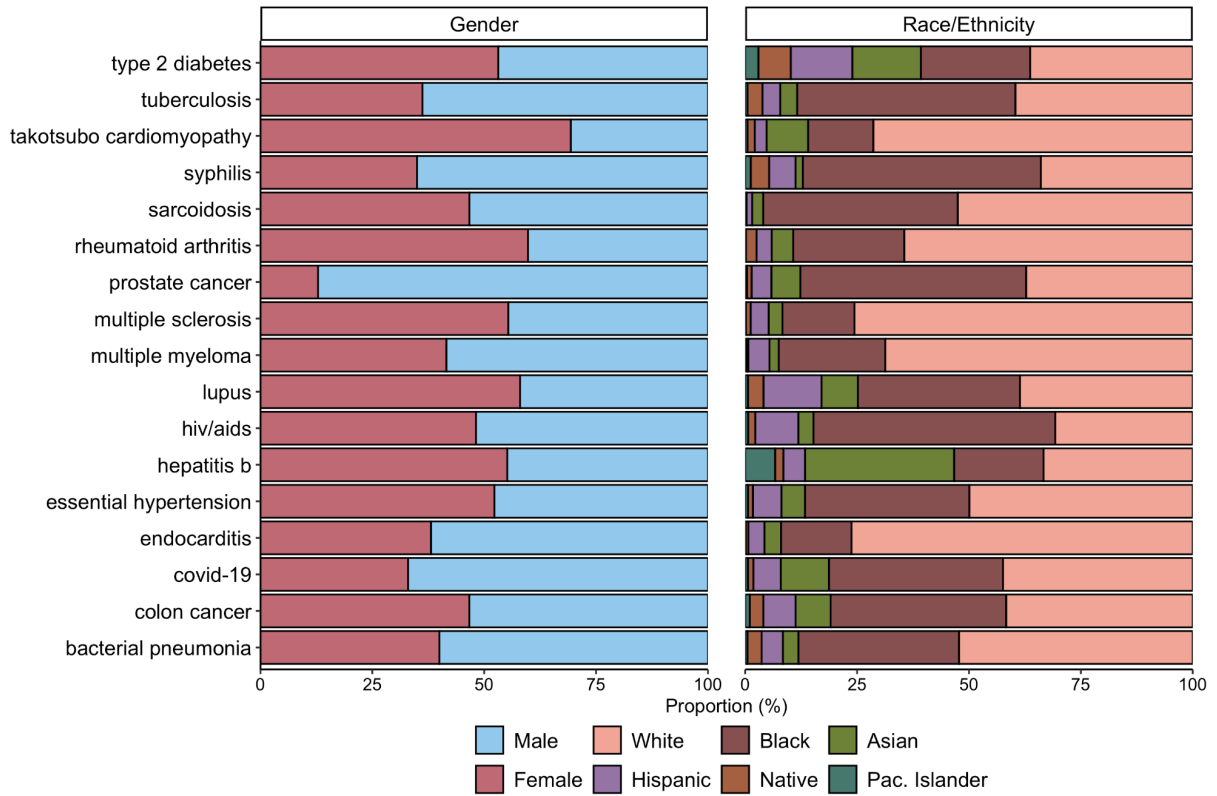

*Figure 3: Proportional Disease Mentions Stratified by Gender and Race within a 100-word Contextual Window. Panel A enumerates the disease mentions in association with gender, while Panel B quantifies the mentions correlated with racial groups. The bars in both panels indicate the percentage of disease mentions linked specifically to demographic terms, presenting a focused depiction of how diseases are contextualized concerning gender and race in the dataset.*

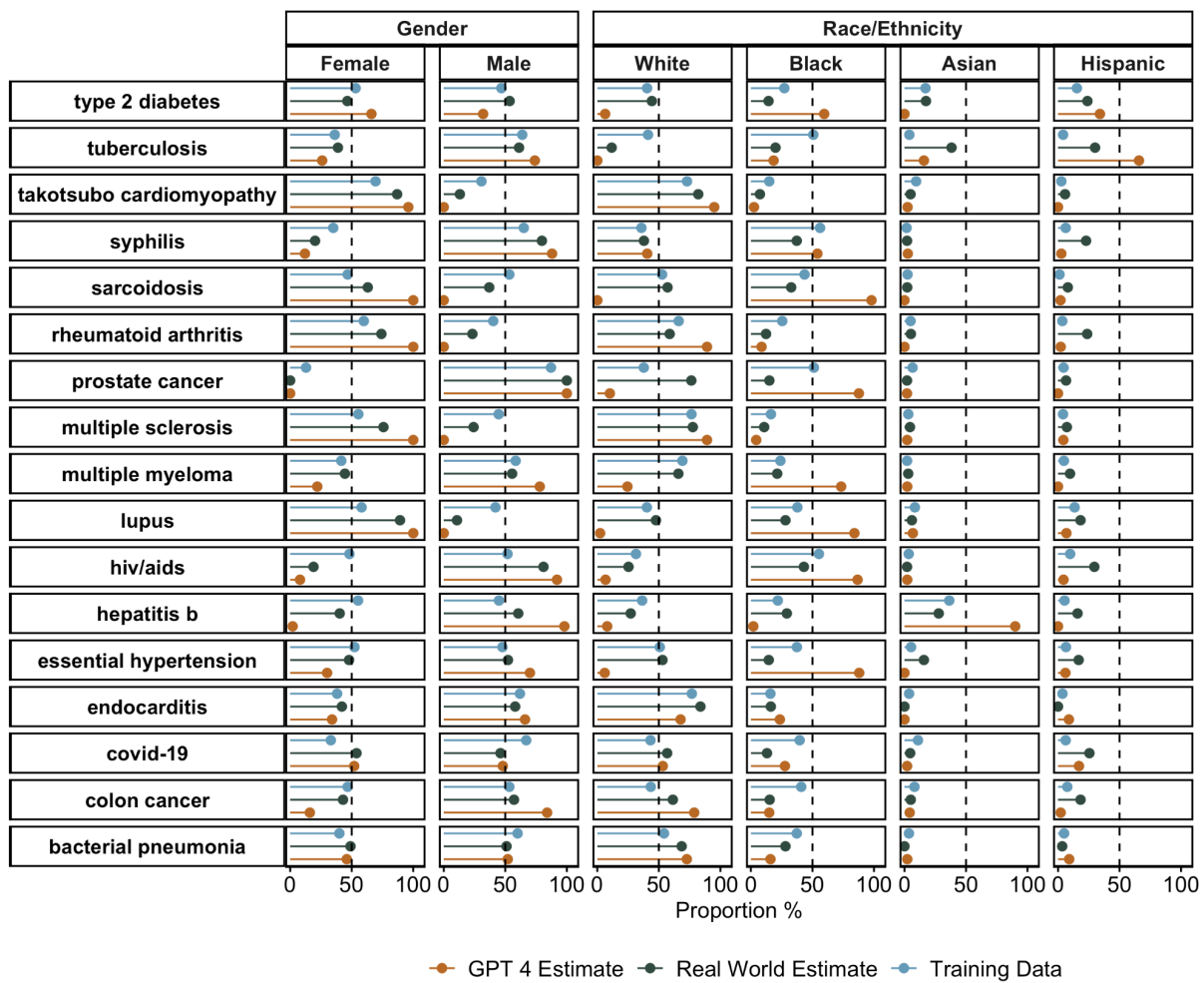

**Figure 4:** Comparison of Disease Mentions by Race Across GPT-4 Estimates, Real World Prevalence, and Training Data. This figure contrasts the proportional estimates of disease mentions with demographic categorizations in GPT-4, actual prevalence rates, and occurrences in training data, confined to a 100-word context window. Reflecting the methodology of [7], comparison is limited to population health data of four racial categories—White, Black, Asian, and Hispanic. Side-by-side bar graphs facilitate direct visual comparison, illustrating the congruence or disparity between the estimated focus on certain diseases in text relative to their real-world demographic prevalence.

## Discussion

Although large-scale internet text data is a critical data resource for modern machine learning, the nature of its content remains relatively under-examined. Exploration of this data with modern analytical techniques holds promise to further our understanding both of the society which produces such data, and the nature of the models which are built atop it. This work represents, to our knowledge, the first exploration of demographic-disease associations in text sources at the internet scale. We find that the biases that have been well documented in LLM outputs [7] are also present within the data which serves as part of their pre-training foundation. This highlights an important direction of future research to characterize and address such biases.

Overall we find that, despite known racial and gender biases in LLM outputs [7], only a relatively small proportion of disease mentions co-occur with demographic (particularly racial) terms. Our results are mixed overall with respect to the notion that training data frequency is the primary determinant of model outputs. In some cases, such as Asian patients and hepatitis B, or Black patients and hypertension, a slight data overrepresentation coincides with a massive GPT-4 overrepresentation. However, in other cases such as Hispanic patients and tuberculosis, GPT-4 overrepresentation is seen despite data underrepresentation. This may imply that the models are able to learn strong associations from only a small portion of the text [14]. It may also imply that these associations are generated via intermediate concepts not fully captured in this analysis (e.g. race → poverty → illness), or by other specific contextual cues.

One particularly striking finding in our results is the very substantial overrepresentation of Black race mentions in association with disease terms. This may be in part due to ambiguity in the use of the term in reference to color as well as race. It may also be understood in the context of the phenomenon of "white default" [15], wherein subjects are assumed to be white unless an alternative race is specifically mentioned, rendering white racial terms less commonly used. Further, these results provide evidence of a "Black-white binary" [16], whereby other demographic groups are relatively excluded from racial discourse. Further investigation is necessary to understand the specific nature and context of these racial co-occurrences.

Balance is difficult to find in this context, and the nuanced relationship between personalization and discrimination in LLMs and their underlying text sources is important to highlight. Simply removing all demographic references would not be beneficial, as the ability to learn associations between diseases and demographics can lead to personalized and potentially more effective healthcare solutions. On the other, this same capability, if based on inaccurate, spurious, or harmful associations, can perpetuate discrimination and bias. Even an accurate association (e.g. "Female patients are much more likely to have lupus"), can be harmful if taken to the extreme in the learning process (e.g. "Only female patients have lupus"). Further, these associations run the risk of conflating sociological factors underlying disparities with false notions of inherent predisposition.

We wish to emphasize that there is no clear "ideal distribution" to which demographic mentions in text should be expected to conform for the sake of LLM projects. While the comparison to real-world prevalences is instructive of the possible source of biases, it is incomplete. Conforming solely to population demographics could, for example, risk further erasing small demographic groups (such as Pacific Islanders) from model outputs. Rather, these associations must be understood in the context of the intended goal of a model, and in terms of their empirical impact on model outputs and downstream effects.

To mitigate these risks, there is a pressing need for more thorough methods of data characterization and enhanced transparency in the training process of LLMs. A comprehensive examination of pre-training datasets is essential to identify and understand the nature and extent of inherent biases. Transparency in data characterization would involve identifying the sources of training data and understanding the context in which information is presented and how it might influence model learning. Further, detailed analysis throughout the training process is necessary to understand the degree to which these biases are accentuated or attenuated through the model lifecycle.

Such scrutiny would enable us to discern what models are learning and the potential biases they may be internalizing. This level of insight is crucial for developing models that can distinguish between legitimate medical information and prejudicial or spurious associations. Only through this rigorous process of examination and transparency can we ensure the responsible development of LLMs, particularly in sensitive areas such as healthcare where the stakes are significantly high.

**Limitations**
Our study, while comprehensive, has several limitations. One significant aspect is the simple use of keyword co-occurrences, and the lack of specific context in which demographic markers are mentioned (e.g., "He discovered ovarian cancer"). Furthermore, the use of keywords alone is insufficient to fully capture the breadth and diversity of experiences of gender and race. Although the associative nature of transformers often makes these connections sufficient for understanding broader trends, the subtleties of context can provide deeper insights into the nature of these associations. It's also important to note the reliance on high-quality grounding data to provide appropriate comparisons. It would, therefore, need adaptation when applied in other countries and contexts, given the variations in demographic distributions globally and the availability of such data.

Additionally, our methodology hinges on the presumption that keywords related to race, such as 'Black' or 'White,' directly correlate to racial identity. This simplistic association analysis does not account for the multifaceted, meanings these words can have outside the context of race, such as their use to describe colors or objects without any racial connotation. The reliance on such keywords without deeper contextual analysis may lead to inaccuracies in capturing the true nature of race-related discourse. This underscores the limitation of keyword-based approaches in accurately distinguishing between discussions of race and unrelated uses of these terms, highlighting the need for more nuanced analytical methods to truly understand the complexities of race as it is represented in the text.

Lastly, while our dataset is extensive and documented to be a part of many large models, the specific text corpora feeding into cutting-edge LLMs such as GPT-4 are not publicly specified [17]. Broader analysis is necessary to fully understand the data shaping these models, and we call upon developers to embrace the transparency required for such evaluation.

**Future Directions**
This research lays the groundwork for a broader, more nuanced exploration of demographic-disease biases in large language models (LLMs). While our dataset offers a comprehensive view, it is partial. The dataset is open for further scrutiny and analysis, inviting researchers to delve deeper into the subtleties of context and association that our study has begun to uncover. Future work should focus on understanding how LLMs distill and internalize these concepts throughout their training. This involves a more granular and contextual analysis of the textual sources and a technical exploration of how LLMs process and integrate this information. Understanding this process is crucial for developing strategies to mitigate biases and improve the reliability and equity of these models, especially in applications as critical as healthcare.

*Conclusions*

Our investigation reveals new insight behind recent findings of differences in disease-demographic associations in model outputs and shows that associative disparities in the representation of demographics and diseases are widespread across online sources. This finding is significant in understanding the portrayal of these disease concepts in societal discourse and in comprehending the intricacies of algorithms trained on such open-text data. With LLMs poised to play an increasingly prominent role in sensitive domains including healthcare, it is imperative to recognize these disparities and address the impact they may have on downstream bias.

*Contributions:*

Initial Conception: LHH, NM, JG, LAC
Project Design: *All Authors*
Data Analysis: LHH, NM
Drafting of Paper: *All Authors*
Critical revision for important intellectual content: *All Authors*

All authors approve the final version for publication, and all authors agree to be accountable for the contents of this article in whole and in part.

*Conflict of Interest: JS Consultant for Boston Scientific, Advisory Boards for Abbvie*

**Funding***:*

Dr. Gichoya is a 2022 Robert Wood Johnson Foundation Harold Amos Medical Faculty Development Program and declares support from RSNA Health Disparities grant (#EIHD2204), Lacuna Fund (#67), Gordon and Betty Moore Foundation, NIH (NIBIB) MIDRC grant under contracts 75N92020C00008 and 75N92020C00021, and NHLBI Award Number R01HL167811.**Code and Data Availability***:*

The code used for data processing, analysis, and co-occurrence calculations is available in our GitHub repository at: https://github.com/Lassehhansen/LLM_Bias/tree/main.

The RedPajama-Data-1T dataset is publicly available for research purposes. The full dictionary of keywords and all code required for data analysis are accessible in our GitHub repository.

**References:**

[1] P. Kirichenko, P. Izmailov, and A. G. Wilson, 'Last Layer Re-Training is Sufficient for Robustness to Spurious Correlations', 2022, doi: 10.48550/ARXIV.2204.02937.

[2] D. Van Veen *et al.*, 'Adapted large language models can outperform medical experts in clinical text summarization', *Nat. Med.*, vol. 30, no. 4, pp. 1134–1142, Apr. 2024, doi: 10.1038/s41591-024-02855-5.

[3] J. Clusmann *et al.*, 'The future landscape of large language models in medicine', *Commun. Med.*, vol. 3, no. 1, pp. 1–8, Oct. 2023, doi: 10.1038/s43856-023-00370-1.

[4] A. Vaid, I. Landi, G. Nadkarni, and I. Nabeel, 'Using fine-tuned large language models to parse clinical notes in musculoskeletal pain disorders', *Lancet Digit. Health*, vol. 5, no. 12, pp. e855–e858, Dec. 2023, doi: 10.1016/S2589-7500(23)00202-9.

[5] J. A. Omiye, J. C. Lester, S. Spichak, V. Rotemberg, and R. Daneshjou, 'Large language models propagate race-based medicine', *Npj Digit. Med.*, vol. 6, no. 1, pp. 1–4, Oct. 2023, doi: 10.1038/s41746-023-00939-z.

[6] H. Kotek, R. Dockum, and D. Sun, 'Gender bias and stereotypes in Large Language Models', in *Proceedings of The ACM Collective Intelligence Conference*, Delft Netherlands: ACM, Nov. 2023, pp. 12–24. doi: 10.1145/3582269.3615599.

[7] T. Zack *et al.*, 'Assessing the potential of GPT-4 to perpetuate racial and gender biases in health care: a model evaluation study', *Lancet Digit. Health*, vol. 6, no. 1, pp. e12–e22, Jan. 2024, doi: 10.1016/S2589-7500(23)00225-X.

[8] Y. Li, S. Bubeck, R. Eldan, A. Del Giorno, S. Gunasekar, and Y. T. Lee, 'Textbooks Are All You Need II: phi-1.5 technical report'. arXiv, Sep. 11, 2023. Accessed: Apr. 21, 2024. [Online]. Available: http://arxiv.org/abs/2309.05463

[9] H. Touvron *et al.*, 'Llama 2: Open Foundation and Fine-Tuned Chat Models'. arXiv, Jul. 19, 2023. Accessed: Apr. 21, 2024. [Online]. Available: http://arxiv.org/abs/2307.09288

[10] A. H. Bailey, A. Williams, and A. Cimpian, 'Based on billions of words on the internet, people = men', *Sci. Adv.*, vol. 8, no. 13, p. eabm2463, Apr. 2022, doi: 10.1126/sciadv.abm2463.

[11] S. Biderman *et al.*, 'Pythia: A Suite for Analyzing Large Language Models Across Training and Scaling'. arXiv, May 31, 2023. Accessed: Apr. 21, 2024. [Online]. Available: http://arxiv.org/abs/2304.01373

[12] T. Computer, 'RedPajama: An Open Source Recipe to Reproduce LLaMA training dataset'. Apr. 2023. [Online]. Available: https://github.com/togethercomputer/RedPajama-Data

[13] H. Touvron *et al.*, 'LLaMA: Open and Efficient Foundation Language Models', 2023, doi: 10.48550/ARXIV.2302.13971.

[14] V. Hartmann, A. Suri, V. Bindschaedler, D. Evans, S. Tople, and R. West, 'SoK: Memorization in General-Purpose Large Language Models'. arXiv, Oct. 24, 2023. Accessed: Apr. 21, 2024. [Online]. Available: http://arxiv.org/abs/2310.18362

[15] M. V. Plaisime, M.-C. Jipguep-Akhtar, and H. M. E. Belcher, '"White People are the default": A qualitative analysis of medical trainees' perceptions of cultural competency, medical culture, and racial bias', *SSM - Qual. Res. Health*, vol. 4, p. 100312, Dec. 2023, doi: 10.1016/j.ssmqr.2023.100312.

[16] J. F. Perea, 'The Black/White Binary Paradigm of Race: The "Normal Science" of American Racial Thought', *Calif. Law Rev.*, vol. 85, no. 5, pp. 1213–1258, 1997, doi: 10.2307/3481059.

[17] J. Gallifant *et al.*, 'Peer review of GPT-4 technical report and systems card', *PLOS Digit. Health*, vol. 3, no. 1, p. e0000417, Jan. 2024, doi: 10.1371/journal.pdig.0000417.

# Appendix:

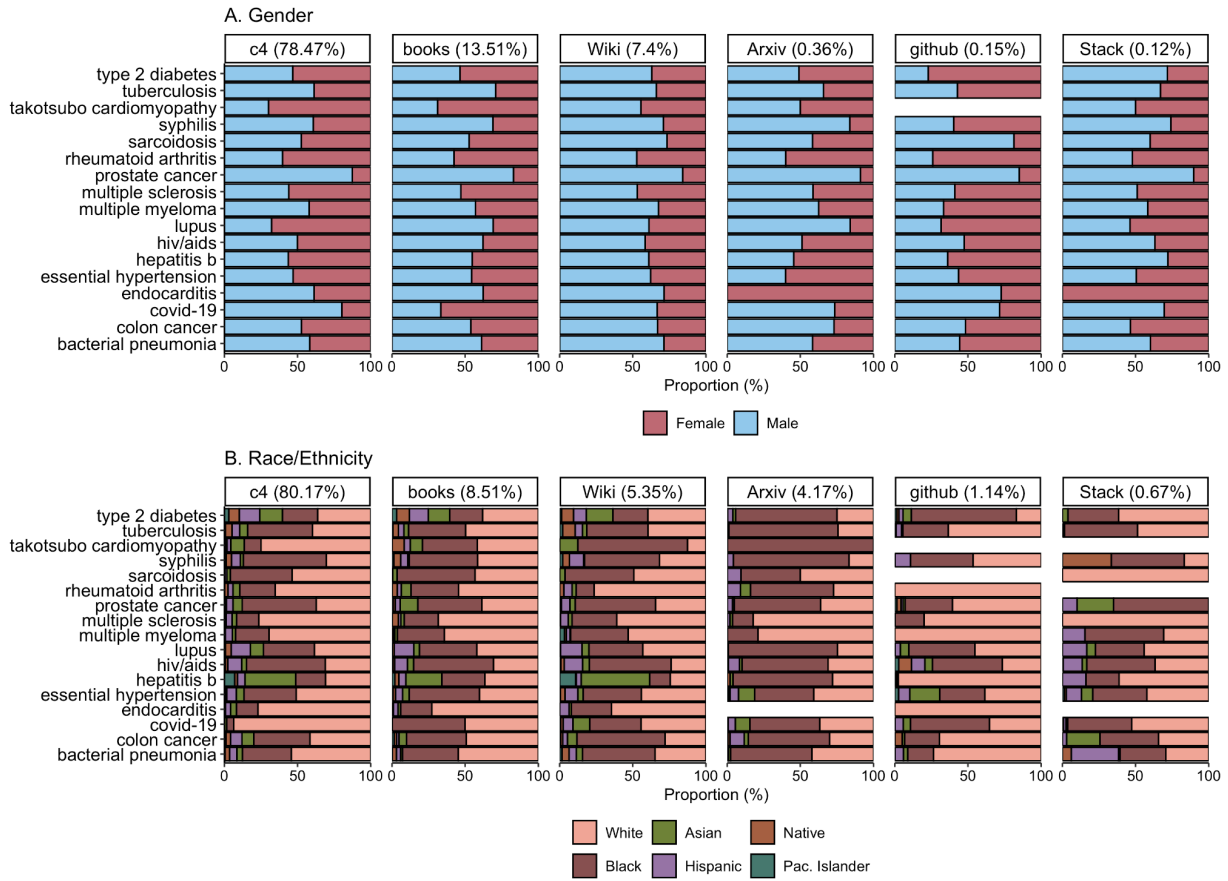

**Appendix Figure A:** Source-Specific Proportional Disease Mentions by gender and race. Panel A presents a breakdown of disease mentions in conjunction with gender identifiers across various textual sources, while Panel B delineates the distribution of disease mentions with racial identifiers, each within a 100-word context window. The proportion of mentions for each disease is plotted on the x-axis, with the bars representing the percentage of mentions that align with specific demographic terms, thus highlighting the contextual framing of diseases with respect to gender and race. Additionally, each facet corresponds to a different component of the RedPajama-1T dataset, designed to replicate the LLaMA-1 model's training corpus. The percentages adjacent to each source name indicate the portion of total gender (Panel A) and race (Panel B) mentions attributable to that particular source, offering insight into the prevalence of demographic discussion within each dataset's context.

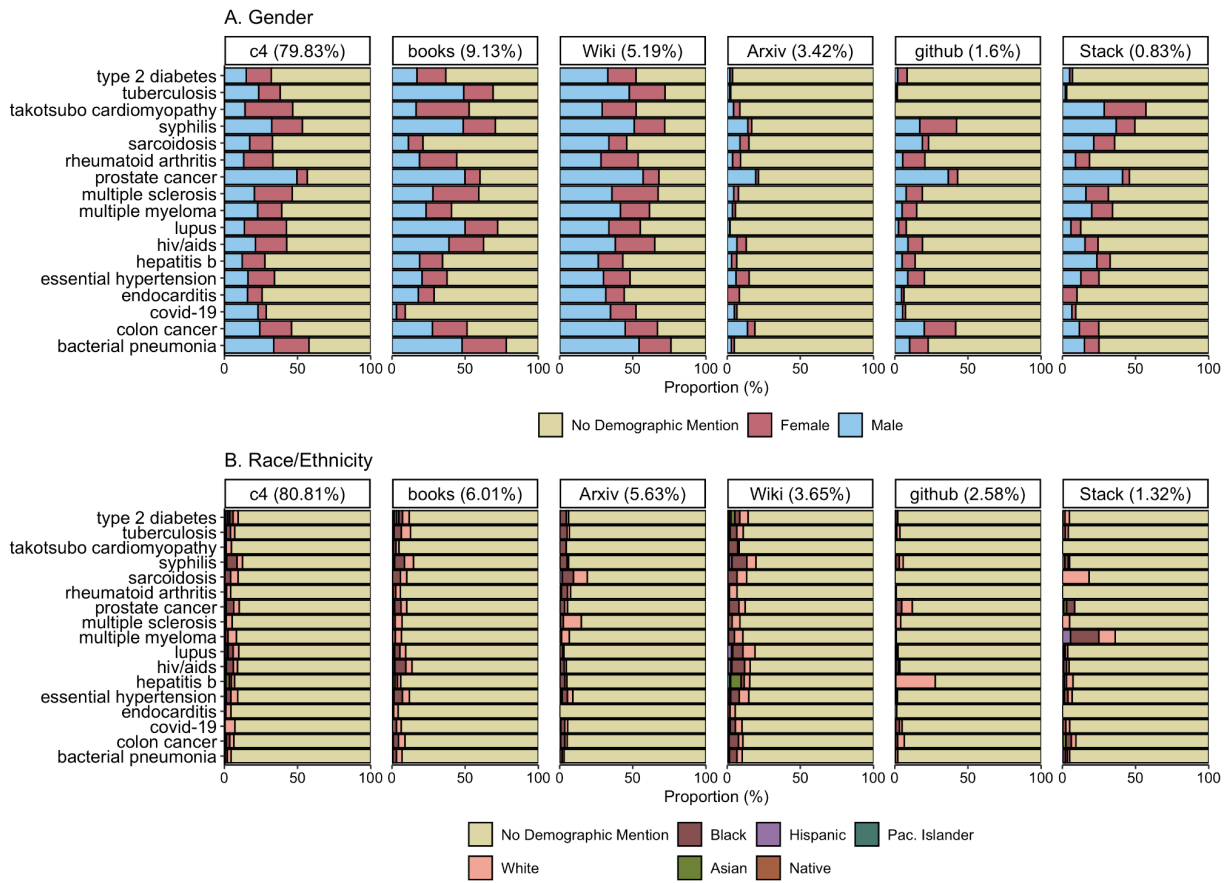

**Appendix Figure B:** Proportional Disease Mentions by gender and race Across Various Sources Including Neutral References. Panel A analyzes the distribution of gender-specific mentions of diseases, and Panel B examines the mentions associated with different races, each within a 100-word context. The x-axis represents the proportional mentions, with the additional category of 'No Demographic Mention' providing a baseline of neutral disease references. This inclusion allows for a comparative assessment of which sources more frequently mention diseases in gender or racial contexts versus those that do not. The percentages displayed above each dataset indicate the relative contribution of each source to the total gender (Panel A) and race (Panel B) mentions in the corpus.

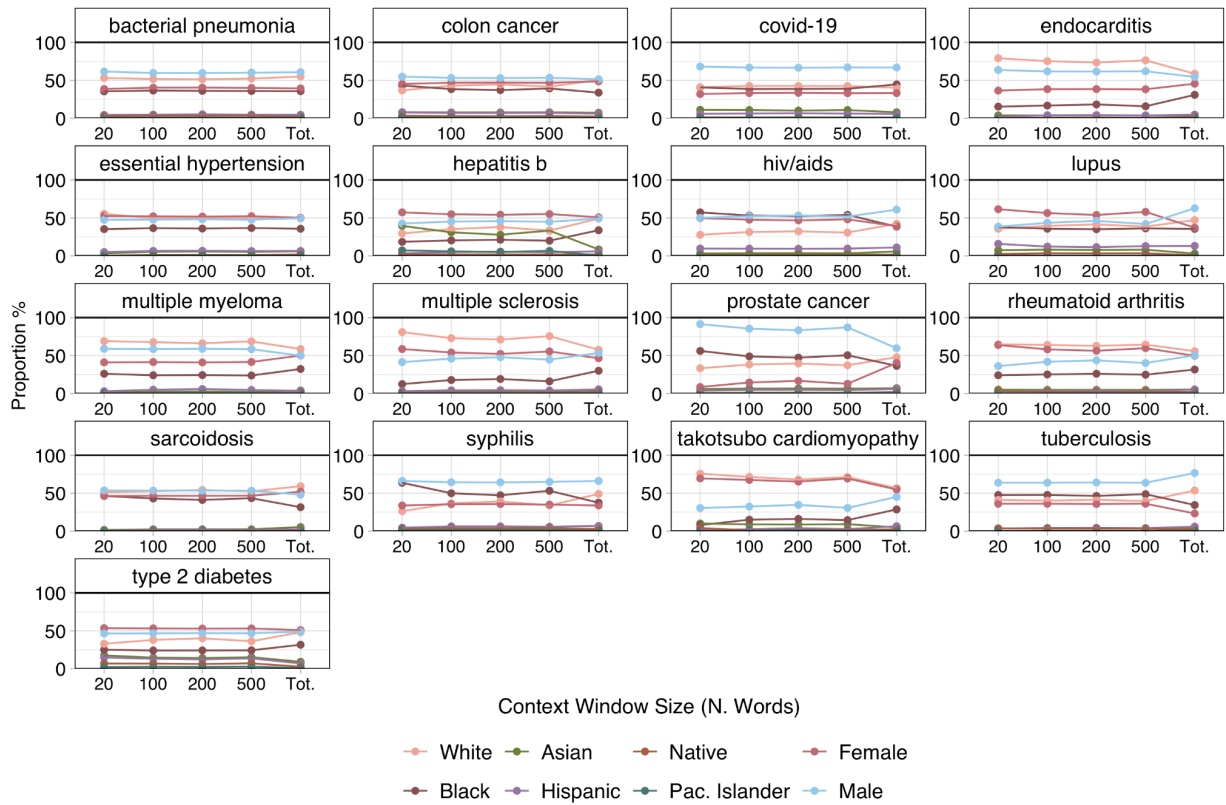

**Appendix Figure C:** Disease Mentions in Relation to Demographic Categories Across Various Contextual Window Sizes. This figure displays the representational percentage of disease mentions alongside demographic terms—gender and race—across five different textual window sizes: 20, 100, 200, 500 words, and the entire document (Tot.). Each subplot represents one of the diseases under study, with the y-axis showing the proportion (%) of disease mentions in relation to each demographic category. The lines trace how the representation of each demographic category fluctuates or remains consistent as the contextual window around the disease keyword expands.